\def\year{2019}\relax
\documentclass[letterpaper]{article} 
\usepackage{aaai19}  
\usepackage{times}  
\usepackage{helvet}  
\usepackage{courier}  
\usepackage{url}  
\usepackage{graphicx}  
\usepackage{amssymb}
\usepackage{array,color}
\usepackage{graphics}
\usepackage{graphicx}
\usepackage{verbatim}
\usepackage{booktabs}
\usepackage{algorithm}
\usepackage{algorithmic}
\usepackage{multirow} 
\usepackage{amsmath}
\usepackage{xcolor}
\usepackage{soul}
\usepackage{subfigure}
\usepackage{caption}
\usepackage[hidelinks]{hyperref}

\DeclareMathOperator{\tr}{tr}

\begin{document}
%
\title{Collaboration based Multi-Label Learning}
\author{Lei Feng$^{1,2}$, Bo An$^1$, Shuo He$^3$\\
$^1$School of Computer Science and Engineering, Nanyang Technological University, Singapore\\
$^2$Alibaba-NTU Singapore Joint Research Institute, Singapore\\
$^3$College of Computer and Information Science, Southwest University, Chongqing, China\\
\{feng0093, boan\}@ntu.edu.sg, hs8207083890@email.swu.edu.cn}
\maketitle
\begin{abstract}
It is well-known that exploiting label correlations is crucially important to multi-label learning. Most of the existing approaches take label correlations as prior knowledge, which may not correctly characterize the real relationships among labels.
Besides, label correlations are normally used to regularize the hypothesis space, while the final predictions are not explicitly correlated.
In this paper, we suggest that \textsl{for each individual label, the final prediction involves the collaboration between its own prediction and the predictions of other labels.}
Based on this assumption, we first propose a novel method to learn the label correlations via sparse reconstruction in the label space.
Then, by seamlessly integrating the learned label correlations into model training, we propose a novel multi-label learning approach that aims to explicitly account for
the correlated predictions of labels while training the desired model simultaneously. Extensive experimental results show that our approach outperforms
the state-of-the-art counterparts.
\end{abstract}
\section{Introduction}
Multi-label learning deals with the problem where an instance can be associated with multiple labels simultaneously. Formally speaking, let $\mathcal{X}\in\mathbb{R}^d$ be $d$-dimensional feature space
 and $\mathcal{Y}=\{y_1, y_2, \cdots, y_q\}$ be the label space with $q$ labels. Given the multi-label training set
 $\mathcal{D}=\{\mathbf{x}_i,\mathbf{y}_i\}_{i=1}^n$ where $\mathbf{x}_i\in\mathcal{X}$ is a feature vector and $\mathbf{y}_i\in\{-1,1\}^q$ is the label vector,
 the goal of multi-label
 learning is to learn a model $f:\mathbb{R}^d\rightarrow \{-1,1\}^q$, which maps from the space of feature vectors to the space of label vectors.
 As a learning framework that handles objects with multiple semantics, multi-label learning has been widely applied in many real-world applications,
 such as image annotation ~\cite{yang2016exploit}, document categorization ~\cite{li2015supervised}, bioinformatics ~\cite{zhang2006multilabel},
 and information retrieval ~\cite{gopal2010multilabel}.

The most straightforward multi-label learning approach~\cite{boutell2004learning} is to decompose the problem into a set of independent binary classification tasks,
one for each label. Although this strategy is easy to implement, it may result in degraded performance, due to the ignorance of correlations among labels. To compensate for this deficiency,
the exploitation of label correlations has been widely accepted as a key component of effective multi-label learning approaches~\cite{gibaja2015tutorial,zhang2014review}.

So far, many methods have been developed to improve the performance of multi-label learning by exploring various types of label
correlations~\cite{tsoumakas2009correlation,cesa2006hierarchical,petterson2011submodular,huang2012multi1,huang2012multi2,zhu2018multi}. There has been increasing interest in
exploiting the label correlations by taking the label correlation matrix as prior knowledge~\cite{hariharan2010large,cai2013new,huang2016learning,huang2018joint}.
Concretely, these methods directly calculate the label correlation matrix by the similarity between label vectors using common similarity measures, and then incorporate the label
correlation matrix into model training for further enhancing the predictions of multiple label assignments. However, the label correlations are simply obtained
by common similarity measures, which may not be able to reflect complex relationships among labels. Besides, these methods exploit label correlations by
manipulating the hypothesis space, while the final predictions are not explicitly correlated.

To address the above limitations, we make a key assumption
that \textsl{for each individual label, the final prediction involves the collaboration between its own prediction and the predictions of other labels}.
Based on this assumption, a novel multi-label learning approach named CAMEL, i.e., CollAboration based Multi-labEl Learning, is proposed.
Different from most of the existing approaches that calculate the label correlation matrix simply by common similarity measures, CAMEL presents a novel method to
learn such matrix and show that it is equivalent to sparse reconstruction in the label space. The learned label correlation matrix is capable of reflecting
the collaborative relationships among labels regarding the final predictions. Subsequently, CAMEL seamlessly incorporates the learned label correlations into
the desired multi-label predictive model. Specifically, label-independent embedding is introduced, which aims to fit the final predictions with the learned
label correlations while guiding the estimation of the model parameters simultaneously. The effectiveness of CAMEL is clearly demonstrated by experimental
results on a number of datasets.

\section{Related Work}
In recent years, many algorithms have been proposed to deal with multi-label learning tasks.
In terms of the $\textsl{order of label correlations}$ being considered, these approaches can be roughly categorized into three strategies~\cite{zhang2014review,gibaja2015tutorial}.

For the first-order strategy, the multi-label learning problem is tackled in a label-by-label manner where label correlations are ignored. Intuitively, one can easily
decompose the multi-label learning problem into a series of independent binary classification problems (one for each label)~\cite{boutell2004learning}. The second-order strategy takes into consideration pairwise relationships between labels, such as the ranking between relevant labels and irrelevant
labels~\cite{elisseeff2002kernel} or the interaction of paired labels~\cite{zhu2005multi}. For the third-order strategy, high-order relationships among labels are considered. Following this strategy, numerous multi-label algorithms are proposed. For example, by modeling all other labels' influences on each label,
a shared subspace~\cite{ji2008extracting} is extracted for model training. By addressing connections among random subsets of labels, a chain of binary classifiers
~\cite{read2011classifier} are sequentially trained.

Recently, there has been increasing interest in second-order approaches~\cite{hariharan2010large,cai2013new,huang2016learning,huang2018joint} that take the label
correlation matrix as prior knowledge for model training. These approaches normally directly calculate the label correlation matrix by the similarity between label vectors using common
similarity measures, and then incorporate the label correlation matrix into model training for further enhancing the predictions of multiple label assignments. For instance,
cosine similarity is widely used to calculate the label correlation matrix~\cite{cai2013new,huang2016learning,huang2018joint}. Such label
correlation matrix is further incorporated into a structured sparsity-inducing norm regularization~\cite{cai2013new} for regularizing the learning hypotheses, or performing joint label-specific feature selection and model training~\cite{huang2016learning,huang2018joint}. In addition, there are also some high-order approaches that exploit label correlations on the hypothesis space, while they
do not rely on the label correlation matrix. For example, a boosting approach~\cite{huang2012multi2} is proposed to exploit label correlations with a hypothesis reuse mechanism.

Note that most of the existing approaches using label correlation matrix are second-order and focus on the hypothesis space. Such simple label correlations exploited in the hypothesis space may not correctly depict the real relationships among labels, and final predictions are not explicitly correlated. In the next section, a
novel high-order approach with crafted label correlation matrix that focus on the label space will be introduced.
\section{The CAMEL Approach}
Following the notations used in Introduction, the training set can be alternatively represented by $\mathcal{D} = \{(\mathbf{X},\mathbf{Y})\}$
where $\mathbf{X}=[\mathbf{x}_1,\mathbf{x}_2,\cdots,\mathbf{x}_n]^\top\in\mathbb{R}^{n\times d}$ denotes the instance matrix,
and $\mathbf{Y} = [\mathbf{y}_1,\mathbf{y}_2,\cdots,\mathbf{y}_n]^\top\in\mathbb{R}^{n\times q}$ denotes the label matrix. In addition, we
denote by $\mathbf{Y}_j\in\mathbb{R}^n$ the $j$-th column vector of the matrix $\mathbf{Y}$ (versus $\mathbf{y}_j\in\mathbb{R}^q$ for the $j$-th row vector of $\mathbf{Y}$),
and $\mathbf{Y}_{-j} = [\mathbf{Y}_1,\cdots,\mathbf{Y}_{j-1},\mathbf{Y}_{j+1},\cdots,\mathbf{Y}_q]\in\mathbb{R}^{n\times (q-1)}$ represents the matrix that
excludes the $j$-th column vector of $\mathbf{Y}$.
\subsection{Label Correlation Learning}
To characterize the collaborative relationships among labels regarding the final predictions, CAMEL works by learning a label correlation
matrix $\mathbf{S} = [s_{ij}]_{q\times q}$ where $s_{ij}$ reflects the contribution of the $i$-label to the $j$-label. Guided by the
assumption that \textsl{for each individual label, the final prediction involves the collaboration between its own prediction and the predictions of other labels}, we
thus take the given label matrix as the final prediction, and propose to learn the label correlation matrix $\mathbf{S}$ in the following way:
\begin{gather}
\label{eq1}
\min_{s_{ij}}\left\|((1-\alpha)\mathbf{Y}_j+\alpha\sum_{i\neq j,i\in [q]}s_{ij}\mathbf{Y}_i) - \mathbf{Y}_j\right\|_2^2
\end{gather}
where $\alpha$ is the tradeoff parameter that controls the collaboration degree. In other words, $\alpha$ is used to balance the $j$-th label's own prediction and the predictions of other labels. Since each label is normally
correlated with only a few labels, the collaborative relationships between one label and other labels could be sparse. With a slight abuse of notation,
we denote by $\mathbf{S}_j = [s_{1j},\cdots,s_{j-1,j},s_{j+1,j},\cdots,s_{qj}]^\top\in\mathbb{R}^{(q-1)}$ the $j$-th column vector
of $\mathbf{S}$ excluding $s_{jj}$ ($s_{jj} = 0$). Under canonical sparse representation, the coefficient vector $\mathbf{S}_j$ is learned by solving
the following optimization problem:
\begin{gather}
\label{eq2}
\min_{\mathbf{S}_j}\left\|(1-\alpha)\mathbf{Y}_{j}+\alpha\mathbf{Y}_{-j}\mathbf{S}_j-\mathbf{Y}_j\right\|_2^2+\hat{\lambda}\left\|\mathbf{S}_j\right\|_1
\end{gather}
where $\hat{\lambda}$ controls the sparsity of the coefficient vector $\mathbf{S}_j$. By properly rewriting the above problem and
setting $\lambda = \hat{\lambda}/{\alpha}$, it is easy to derive the following equivalent optimization problem:
\begin{gather}
\label{eq3}
\min_{\mathbf{S}_j}\left\|\mathbf{Y}_{-j}\mathbf{S}_j-\mathbf{Y}_j\right\|_2^2+\lambda\left\|\mathbf{S}_j\right\|_1
\end{gather}
Here, this problem aims to estimate the collaborative relationships between the $j$-th label and the other labels via sparse reconstruction. The first term corresponds to
the linear reconstruction error via $\ell_2$ norm, and the second term controls the sparsity of the reconstruction coefficients by using $\ell_1$ norm. The relative importance
of each term is balanced by the tradeoff parameter $\lambda$, which is empirically set to $\frac{1}{100}\left\|\mathbf{Y}_{j}^\top\mathbf{Y}_{-j}\right\|_{\infty}$ in the experiments. To solve problem~(\ref{eq3}), the popular Alternating Direction Method of
Multiplier (ADMM)~\cite{boyd2011distributed} is employed, and detailed information is given in Appendix A. After solving problem~(\ref{eq3}) for
each label, the weight matrix $\mathbf{S}$ can be accordingly constructed with all diagonal elements set to 0. Note that for most of the existing second-order approaches
using label correlation matrix~\cite{hariharan2010large,cai2013new,huang2016learning,huang2018joint}, only pairwise relationships are considered, and the relationships between one label and the other labels are separated. While for CAMEL, since the final prediction of each label is determined by all the predictions of other labels and itself, the relationships among all labels are exploited in a collaborative manner. Which means, the relationships between one label and the other labels are coordinated (influenced by each other).
Therefore, CAMEL is a high-order approach.
\subsection{Multi-Label Classifier Training}
In this section, we propose a novel multi-label learning approach by seamlessly integrating the learned label correlations into the desired predictive model.
Suppose the ordinary prediction matrix of $\mathbf{X}$ is denoted by $f(\mathbf{X}) = [f_1(\mathbf{X}), f_2(\mathbf{X}),\cdots, f_q(\mathbf{X})]\in\mathbb{R}^{n\times q}$
where $f_1(\cdot),f_2(\cdot),\cdots,f_q(\cdot)$ denotes the individual label predictors respectively. In the ordinary setting, each label predictor is only in charge of a single
label, while label correlations are fully lost. To absorb the learned label correlations into predictions, we reuse the assumption
that \textsl{for each individual label, the final prediction involves the collaboration between its own prediction and the predictions of other labels}, and propose to
compute the final prediction of the $j$-th label as follows:
\begin{gather}
\label{eq4}
(1-\alpha)f_j(\mathbf{X}) + \alpha\sum_{i\neq j,i\in[q]}s_{ij}f_i(\mathbf{X})
\end{gather}
where $\alpha$ is consistent with problem (\ref{eq1}), which controls the collaboration degree of label predictions. By considering all the $q$ label predictions simultaneously, we thus obtain the following compact representation:
\begin{gather}
\label{eq5}
(1-\alpha)f(\mathbf{X}) + \alpha f(\mathbf{X})\mathbf{S}=f(\mathbf{X})((1-\alpha)\mathbf{I}+\alpha\mathbf{S})
\end{gather}
Here, the whole multi-label learning problem could be considered as two parallel subproblems, i.e., training the ordinary model and fitting the final predictions by the modeling outputs
with label correlations. Thus, we propose to learn label-independent embedding denoted
by $\mathbf{Z}\in\mathbb{R}^{n\times q}$, which works as a bridge between model training and prediction fitting. This brings several advantages: First, the two subproblems can be
solved via alternation, which encourages the mutual adaption of model training and prediction fitting; Second, the relative importance of the two subproblems can be controlled by
a tradeoff parameter; Third, closed-form solutions and kernel extension can be easily derived. Let $\mathbf{G}=(1-\alpha)\mathbf{I}+\alpha\mathbf{S}$,
the proposed formulation is given as follows:
\begin{gather}
\label{eq6}
\min_{\mathbf{Z},f}\frac{1}{2}\left\|f(\mathbf{X})-\mathbf{Z}\right\|_F^2 + \frac{\lambda_1}{2}\left\|\mathbf{Z}\mathbf{G}-\mathbf{Y}\right\|_F^2+\frac{\lambda_2}{2}\Omega(f)
\end{gather}
where $\Omega(f)$ controls the complexity of the model $f$, $\lambda_1$ and $\lambda_2$ are the tradeoff parameters determining the relative importance of the above three terms.
To instantiate the above formulation, we choose to train the widely-used
model $f(\mathbf{X}) = \phi(\mathbf{X})\mathbf{W}+\mathbf{1}\mathbf{b}^\top$ where $\mathbf{W}$ and $\mathbf{b}$ are the model parameters, $\mathbf{1} = [1,\cdots,1]^\top$ denotes the column vector with all elements equal to 1, and $\phi(\mathbf{\cdot})$ is a feature mapping that maps the feature
space to some higher (maybe infinite) dimensional Hilbert space.
For the regularization term to control the model complexity, we adopt the widely-used squared Frobenius norm, i.e., $\left\|\mathbf{W}\right\|_F^2$.
To further facilitate a kernel extension for the general nonlinear case, we finally present the formulation as a constrained optimization problem:
\begin{gather}
\label{eq7}
\min_{\mathbf{W},\mathbf{Z},\mathbf{E},\mathbf{b}}\frac{1}{2}\left\|\mathbf{E}\right\|_F^2+
\frac{\lambda_1}{2}\left\|\mathbf{Z}\mathbf{G}-\mathbf{Y}\right\|_F^2+\frac{\lambda_2}{2}\left\|\mathbf{W}\right\|_F^2\\
\nonumber
\text{s.t.}\quad \mathbf{Z}-\phi(\mathbf{X})\mathbf{W}-\mathbf{1}\mathbf{b}^\top=\mathbf{E}
\end{gather}
\begin{algorithm}[t]
\caption{The CAMEL Algorithm}
\label{alg1}
\begin{algorithmic}[1]
\REQUIRE\mbox{}\par
$\mathcal{D}$: the multi-label training set $\mathcal{D}=\{(\mathbf{X}, \mathbf{Y})\}$\\
$\alpha, \lambda_1, \lambda_2$: the hyperparameters\\
$\mathbf{x}$: the unseen test instance
\ENSURE \mbox{}\par
$\mathbf{y}$: the predicted label for the test instance $\mathbf{x}$\\
\item[]
\STATE learn the label correlation matrix $\mathbf{S}$ by solving problem (\ref{eq3}) for each label via ADMM procedure;
\STATE set $\mathbf{G} = (1-\alpha)\mathbf{I} + \alpha\mathbf{S}$;
\STATE initialize $\mathbf{Z} = \mathbf{Y}$;
\STATE construct the kernel matrix $\mathbf{K}=[\mathcal{K}(\mathbf{x}_i,\mathbf{x}_j)]_{n\times n}$ by Gaussian kernel function;
\REPEAT
\STATE update $\mathbf{b}$ and $\mathbf{A}$ according to (\ref{eq9});
\STATE update $\mathbf{T} = \frac{1}{\lambda_2}\mathbf{KA} + \mathbf{1}\mathbf{b}^\top$;
\STATE update $\mathbf{Z}$ in terms of (\ref{eq11});
\UNTIL convergence.\STATE return the predicted label vector $\mathbf{y}$ according to (\ref{eq12}).
\end{algorithmic}
\end{algorithm}
\section{Optimization}
Problem (\ref{eq7}) is convex with respect to $\mathbf{W}$ and $\mathbf{b}$ with $\mathbf{Z}$ fixed, and also convex with respect to $\mathbf{Z}$ with $\mathbf{W}$ and $\mathbf{b}$ fixed. Therefore, it is a biconvex problem~\cite{gorski2007biconvex}, and can be solved by an alternating approach.
\subsubsection{Updating $\mathbf{W}$ and $\mathbf{b}$ with $\mathbf{Z}$ fixed}
With $\mathbf{Z}$ fixed, problem~(\ref{eq7}) reduces to
\begin{gather}
\label{updateW}
\min_{\mathbf{E},\mathbf{W},\mathbf{b}}\frac{1}{2}\left\|\mathbf{E}\right\|_F^2+\frac{\lambda_2}{2}\left\|\mathbf{W}\right\|_F^2\\
\nonumber
\text{s.t.}\quad \mathbf{Z}-\phi(\mathbf{X})\mathbf{W}-\mathbf{1}\mathbf{b}^\top=\mathbf{E}
\end{gather}
By deriving the Lagrangian of the above constrained problem and setting the gradient with respect to $\mathbf{W}$ to 0,
it is easy to show $\mathbf{W}=\frac{1}{\lambda_2}\phi(\mathbf{X})^\top\mathbf{A}$ where $\mathbf{A}=[a_{ij}]_{n\times q}$ is the matrix that stores the Lagrangian multipliers.
Let $\mathbf{K}=\phi(\mathbf{X})\phi(\mathbf{X})^\top$ be the kernel matrix with its element
$k_{ij} = \mathcal{K}(\mathbf{x}_i,\mathbf{x}_j)=\phi(\mathbf{x}_i)^\top\phi(\mathbf{x}_j)$, where $\mathcal{K}(\cdot,\cdot)$ represents the kernel function.
For CAMEL, Gaussian kernel function $\mathcal{K}(\mathbf{x}_i,\mathbf{x}_j) = \exp(-\left\|\mathbf{x}_i-\mathbf{x}_j\right\|_2^2/(2\sigma^2))$ is employed with $\sigma$ set to the average Euclidean distance of all pairs of training instances.
In this way, we choose to optimize with respect to $\mathbf{A}$ and $\mathbf{b}$ instead, and the close-form solutions are reported as follows:
\begin{align}
\label{eq9}
\mathbf{b}^\top &= \frac{\mathbf{1}^\top\mathbf{H}^{-1}\mathbf{Z}}{\mathbf{1}^\top\mathbf{H}^{-1}\mathbf{1}}\\
\nonumber
\mathbf{A} &= \mathbf{H}^{-1}(\mathbf{Z}-\mathbf{1}\mathbf{b}^\top)
\end{align}
where $\mathbf{H} = \frac{1}{\lambda_2}\mathbf{K}+\mathbf{I}$. The detailed information is provided in Appendix B.
\subsubsection{Updating $\mathbf{Z}$ with $\mathbf{W}$ and $\mathbf{b}$ fixed}
When $\mathbf{W}$ and $\mathbf{b}$ are fixed, the modeling output matrix $\mathbf{T}\in\mathbb{R}^{n\times q}$ is calculated by
$\mathbf{T} = \phi(\mathbf{X})\mathbf{W}+\mathbf{1}\mathbf{b}^\top = \frac{1}{\lambda_2}\mathbf{K}\mathbf{A}+\mathbf{1}\mathbf{b}^\top$. By inserting $\mathbf{E}=\mathbf{Z}-\mathbf{T}$, problem~(\ref{eq7}) reduces to:
\begin{gather}
\label{eq10}
\min_{\mathbf{Z}}\frac{1}{2}\left\|\mathbf{Z}-\mathbf{T}\right\|_F^2 + \frac{\lambda_1}{2}\left\|\mathbf{Z}\mathbf{G}-\mathbf{Y}\right\|_F^2
\end{gather}
Setting the gradient with respect to $\mathbf{Z}$ to 0, we can obtain the following closed-form solution:
\begin{gather}
\label{eq11}
\mathbf{Z} = (\mathbf{T}+\lambda_1\mathbf{Y}\mathbf{G}^\top)(\mathbf{I}+\lambda_1\mathbf{G}\mathbf{G}^\top)^{-1}
\end{gather}

Once the iterative process converges, the predicted label vector $\mathbf{y}\in\{-1,1\}^l$ of the test instance $\mathbf{x}$ is given as:
\begin{gather}
\label{eq12}
\mathbf{y} = \text{sign}(\mathbf{G}^\top(\sum_{i=1}^m\mathbf{a}_{i}\mathcal{K}(\mathbf{x},\mathbf{x}_i)+\mathbf{b}))
\end{gather}
The pseudo code of CAMEL is presented in Algorithm~\ref{alg1}. Since the proposed formulation is biconvex, this alternating optimization process is
guaranteed to converge~\cite{gorski2007biconvex}.
\section{Experiments}
In this section, we conduct extensive experiments on various datasets to validate the effectiveness of CAMEL.
\subsection{Experimental Setup}
\subsubsection{Datasets}
For comprehensive performance evaluation, we collect sixteen benchmark multi-label datasets. 
For each dataset $\mathcal{S}$, we denote by $|\mathcal{S}|$, $dim(\mathcal{S})$, $L(\mathcal{S})$, $LCard(\mathcal{S})$, and $F(\mathcal{S})$ the number of examples, the number of features (dimensions), the number of distinct class labels, the average number of labels associated with each example, and feature type, respectively. Table 1 summarizes the detailed characteristics of these datasets, which are organized in ascending order of $|\mathcal{S}|$. According to $|\mathcal{S}|$, we further roughly divide these datasets into regular-size datasets ($|\mathcal{S}|<5000$) and large-size datasets ($|\mathcal{S}|\geq 5000$). For performance evaluation, 10-fold cross-validation is conducted on these datasets, where mean metric values with standard deviations are recorded.
\subsubsection{Evaluation Metrics}
For performance evaluation, we use seven widely-used evaluation metrics, including \textsl{One-error}, \textsl{Hamming loss}, \textsl{Coverage}, \textsl{Ranking loss}, \textsl{Average precision}, \textsl{Macro-averaging F1}, and \textsl{Micro-averaging F1}. Note that for all the employed multi-label evaluation metrics, their values vary within the interval [0,1].
In addition, for the last three metrics, the larger values indicate the better performance, and we use the symbol $\uparrow$ to present such positive logic. While for the first five metrics, the smaller values indicate the better performance, which is represented by $\downarrow$. More detailed information about these evaluation metrics can be found in~\cite{zhang2014review}.
\subsubsection{Comparing Approaches}
CAMEL is compared with three well-established and two state-of-the-art multi-label learning algorithms, including the first-order approach BR~\cite{boutell2004learning}, the second-order approaches LLSF~\cite{huang2016learning} and JFSC~\cite{huang2018joint}, and the high-order approaches ECC~\cite{read2011classifier}, and RAKEL~\cite{tsoumakas2011random}. Here, LLSF and JFSC are the state-of-the-art counterparts using label correlation matrix.

BR, ECC, and RAKEL are implemented under the MULAN multi-label learning package~\cite{tsoumakas2011mulan} by using the logistic regression model as the base classifier. Furthermore, parameters suggested in the corresponding literatures are used, i.e., ECC: ensemble size 30; RAKEL: ensemble size $2q$ with $k=3$. For LLSF, parameters $\alpha,\beta$ are chosen from $\{2^{-10},2^{-9},\cdots,2^{10}\}$, and $\rho$ chosen from $\{0.1, 1, 10\}$. For JFSC, parameters $\alpha,\beta$, and $\gamma$ are chosen from $\{4^{-5},4^{-4},\cdots,4^5\}$, and $\eta$ is chosen from $\{0.1,1,10\}$. For the proposed approach CAMEL, 
$\lambda_1$ is empirically set to 1, $\lambda_2$ is chosen from $\{10^{-3},2\times10^{-3},10^{-2},2\times10^{-2},\cdots,10^0\}$, and $\alpha$ is chosen from $\{0,0.1,\cdots,1\}$. All of these parameters are decided by conducting 5-fold cross-validation on training set.
\begin{table}[htbp]
	\caption{Characteristics of the benchmark multi-label datasets.}
    \setlength{\tabcolsep}{1.5mm}
	\centering
	\begin{tabular}{lccccc}
		\hline
		\hline
		Dataset & $|\mathcal{S}|$    & $dim(\mathcal{S})$ & $L(\mathcal{S})$ & $LCard(\mathcal{S})$  & $F(\mathcal{S})$  \\
		\hline
		cal500   & 502  & 68     & 174  & 26.04   & numeric\\    
		emotions & 593  & 72     & 6    & 1.87    & numeric  \\    
		genbase  & 662  & 1185   & 27   & 1.25    & nominal\\    
		medical  & 978  & 1449   & 45   & 1.25    & nominal\\    
		enron    & 1702 & 1001   & 53   & 3.38    & nominal\\    
		image    & 2000 & 294    & 5    & 1.24    & numeric \\   
		scene    & 2407 & 294    & 5    & 1.07    & numeric \\ 
		yeast    & 2417 & 103    & 14   & 4.24    & numeric \\    
		\hline
		science  & 5000 & 743    & 40   & 1.45    & numeric \\    
		arts     & 5000 & 462    & 26   & 1.64    & numeric \\    
        business & 5000 & 438    & 30   & 1.59    & numeric \\
		rcv1-s1  & 6000 & 944    & 101  & 2.88    & nominal \\    
		rcv1-s2  & 6000 & 944    & 101  & 2.63    & nominal \\    
		rcv1-s3  & 6000 & 944    & 101  & 2.61    & nominal \\    
		rcv1-s4  & 6000 & 944    & 101  & 2.48    & nominal \\    
		rcv1-s5  & 6000 & 944    & 101  & 2.64    & nominal \\    
		\hline
		\hline
	\end{tabular}
	\label{dataset}
\end{table}
\subsection{Experimental Results}
	\begin{table*}[!t]
		\centering
		\caption{Predictive performance of each algorithm (mean$\pm$std. deviation) on the regular-scale datasets.}
		\label{regularResult}
        \resizebox{\textwidth}{85mm}{
		\begin{tabular}{ccccccccc}
			\hline
			\hline
			\multicolumn{1}{c}{\multirow{2}{*}{\begin{tabular}[c]{@{}c@{}}Comparing\\ algorithms\end{tabular}}} & \multicolumn{8}{c}{One-error$\downarrow$}\\ \cline{2-9}
			\multicolumn{1}{c}{}    & cal500 & emotions  & genbase  & medical   & enron  & image  & scene   & yeast \\
			\hline
			CAMEL &0.129$\pm$0.053 &0.292$\pm$0.052 &\textbf{0.001$\pm$0.001} & \textbf{0.110$\pm$0.021} & \textbf{0.207$\pm$0.038}&\textbf{0.242$\pm$0.033}  &\textbf{0.175$\pm$0.027}  &\textbf{0.218$\pm$0.027}\\
			BR &0.893$\pm$0.038   &\textbf{0.284$\pm$0.077}    &0.017$\pm$0.016 	    &0.322$\pm$0.055  		 &0.646$\pm$0.023          &0.387$\pm$0.027  &0.361$\pm$0.036  &0.244$\pm$0.028  \\
			ECC  & 0.295$\pm$0.036 	&0.296$\pm$0.074  	   &0.010$\pm$0.013  		&0.156$\pm$0.037  		 &0.421$\pm$0.034          &0.406$\pm$0.023  &0.306$\pm$0.020  &0.238$\pm$0.030  \\
			RAKEL   &0.634$\pm$0.039   &0.300$\pm$0.070  	   &0.009$\pm$0.007 		&0.243$\pm$0.055  		 &0.532$\pm$0.007   &0.402$\pm$0.024  &0.280$\pm$0.031  &0.244$\pm$0.027  \\
			LLSF  &0.138$\pm$0.050   &0.412$\pm$0.051 & 0.002$\pm$0.005  		&0.120$\pm$0.020  		 &0.250$\pm$0.042          &0.327$\pm$0.030  &0.259$\pm$0.020  &0.394$\pm$0.029  \\
			JFSC & \textbf{0.116$\pm$0.051}   &0.438$\pm$0.086  	   &0.002$\pm$0.005  		&0.128$\pm$0.024         &0.278$\pm$0.041          &0.346$\pm$0.023  &0.266$\pm$0.022  &0.242$\pm$0.021  \\
            \hline
			\multicolumn{1}{c}{\multirow{2}{*}{\begin{tabular}[c]{@{}c@{}}Comparing\\ algorithms\end{tabular}}} & \multicolumn{8}{c}{Hamming loss$\downarrow$}\\ \cline{2-9}
			\multicolumn{1}{c}{}    & cal500 & emotions  & genbase  & medical   & enron  & image  & scene   & yeast \\
			\hline
			CAMEL &\textbf{0.136$\pm$0.005} & \textbf{0.203$\pm$0.021} & \textbf{0.001$\pm$0.001}&0.011$\pm$0.001  & \textbf{0.045$\pm$0.003} & \textbf{0.144$\pm$0.012} &\textbf{0.072$\pm$0.009} & \textbf{0.190$\pm$0.005}\\
			BR &0.189$\pm$0.005  &0.216$\pm$0.028  			   & 0.002$\pm$0.002 & 0.026$\pm$0.003   & 0.111$\pm$0.006 & 0.210$\pm$0.014   &0.139$\pm$0.009  &0.205$\pm$0.007  \\
			ECC  &0.154$\pm$0.005  &0.214$\pm$0.027   & 0.009$\pm$0.004  		&0.011$\pm$0.002 &0.067$\pm$0.002 		&0.210$\pm$0.016  		 &0.112$\pm$0.006 &0.204$\pm$0.010  \\
			RAKEL   &0.195$\pm$0.004  &0.238$\pm$0.025     &0.002$\pm$0.001  		&0.020$\pm$0.002  		   &0.092$\pm$0.004  		&0.223$\pm$0.013  		 &0.139$\pm$0.008 &0.224$\pm$0.009  \\
			LLSF  &0.138$\pm$0.006  &0.267$\pm$0.022  	   &\textbf{0.001$\pm$0.001}  		&\textbf{0.010$\pm$0.002} 		   &0.048$\pm$0.002  		& 0.180$\pm$0.010  	 	 &0.109$\pm$0.003  & 0.278$\pm$0.009  \\
		    JFSC & 0.191$\pm$0.004  &0.295$\pm$0.019     &\textbf{0.001$\pm$0.001}  		&\textbf{0.010$\pm$0.001} 	   &0.051$\pm$0.003  		&0.188$\pm$0.012  		 &0.110$\pm$0.007  &0.206$\pm$0.006  \\
			\hline
			\multicolumn{1}{c}{\multirow{2}{*}{\begin{tabular}[c]{@{}c@{}}Comparing\\ algorithms\end{tabular}}} & \multicolumn{8}{c}{Coverage$\downarrow$}\\ \cline{2-9}
			\multicolumn{1}{c}{}    & cal500 & emotions  & genbase  & medical   & enron  & image  & scene   & yeast \\
			\hline
			CAMEL &0.752$\pm$0.019 &0.312$\pm$0.031 &\textbf{0.012$\pm$0.005}  	   &\textbf{0.028$\pm$0.012} 	&\textbf{0.239$\pm$0.028} 	 &\textbf{0.156$\pm$0.016}  & \textbf{0.062$\pm$0.006} & \textbf{0.446$\pm$0.010}  \\
			BR &0.786$\pm$0.015 & 0.319$\pm$0.026  	&0.014$\pm$0.006 & 0.113$\pm$0.030 & 0.580$\pm$0.023 & 0.216$\pm$0.018    & 0.168$\pm$0.015 & 0.463$\pm$0.011  \\
			ECC  &0.796$\pm$0.019  			&\textbf{0.310$\pm$0.029} 		  &0.013$\pm$0.003 	   &0.034$\pm$0.012 & 0.291$\pm$0.020  		 &0.233$\pm$0.022 & 0.135$\pm$0.010  &0.460$\pm$0.010  \\
			RAKEL   &0.962$\pm$0.016  	&0.362$\pm$0.027    &0.014$\pm$0.005  	   &0.095$\pm$0.018  		&0.513$\pm$0.019  		 &0.253$\pm$0.017  		  &0.169$\pm$0.013  &0.544$\pm$0.013  \\
			LLSF  &0.778$\pm$0.025  	&0.362$\pm$0.032  &0.021$\pm$0.006     	   &0.031$\pm$0.014  		&0.283$\pm$0.023  		 &0.192$\pm$0.007  		  &0.092$\pm$0.006  &0.601$\pm$0.020  \\
			JFSC &\textbf{0.730$\pm$0.026}  &0.392$\pm$0.046    &0.014$\pm$0.007  	   &0.030$\pm$0.012  		&0.314$\pm$0.024  		 &0.200$\pm$0.009  		  &0.102$\pm$0.007 &0.455$\pm$0.011  \\
			\hline
			\multicolumn{1}{c}{\multirow{2}{*}{\begin{tabular}[c]{@{}c@{}}Comparing\\ algorithms\end{tabular}}} & \multicolumn{8}{c}{Ranking loss$\downarrow$}\\ \cline{2-9}
			\multicolumn{1}{c}{}    & cal500 & emotions  & genbase  & medical   & enron  & image  & scene   & yeast \\
			\hline
			CAMEL &\textbf{0.177$\pm$0.009 } & 0.180$\pm$0.032 & \textbf{0.001$\pm$0.001}  	&\textbf{0.016$\pm$0.008 } & \textbf{0.079 $\pm$0.028}  		&\textbf{0.128$\pm$0.013} & \textbf{0.058$\pm$0.005 } & \textbf{0.162$\pm$0.007}  \\
			BR &0.233$\pm$0.007  &0.182$\pm$0.030  		& 0.003$\pm$0.004 & 0.091$\pm$0.027  		   &0.304$\pm$0.014 & 0.204$\pm$0.017  		 &0.151$\pm$0.015  		   &0.176$\pm$0.008  \\
			ECC  &0.219$\pm$0.007  &\textbf{0.172$\pm$0.031}  		&0.002$\pm$0.002  		 &0.022$\pm$0.010   	   &0.118$\pm$0.008 & 0.225$\pm$0.023  		 &0.117$\pm$0.010  		   &0.179$\pm$0.009  \\
		    RAKEL   &0.366$\pm$0.008  &0.225$\pm$0.029  	&0.002$\pm$0.001  		 &0.073$\pm$0.018 		   &0.244$\pm$0.017  		&0.221$\pm$0.018  		 &0.131$\pm$0.014 		   &0.240$\pm$0.009  \\
			LLSF  &0.184$\pm$0.012  &0.245$\pm$0.033 & 0.002$\pm$0.003  		 &0.019$\pm$0.010  		   &0.107$\pm$0.009  		&0.174$\pm$0.006  		 &0.093$\pm$0.005  		   &0.346$\pm$0.017  \\
			JFSC &0.188$\pm$0.010  &0.271$\pm$0.041	&0.003$\pm$0.003  		 &0.017$\pm$0.008  		   &0.118$\pm$0.013  		&0.183$\pm$0.007  		 &0.105$\pm$0.007  		   &0.179$\pm$0.009  \\
			\hline
			\multicolumn{1}{c}{\multirow{2}{*}{\begin{tabular}[c]{@{}c@{}}Comparing\\ algorithms\end{tabular}}} & \multicolumn{8}{c}{Average precision$\uparrow$}\\ \cline{2-9}
			\multicolumn{1}{c}{}    & cal500 & emotions  & genbase  & medical   & enron  & image  & scene   & yeast \\
			\hline
			CAMEL &\textbf{0.515$\pm$0.018} & 0.788$\pm$0.035  &\textbf{0.997$\pm$0.003}  &\textbf{0.917$\pm$0.017} &\textbf{0.718$\pm$0.025} &\textbf{0.843$\pm$0.018}   &\textbf{0.897$\pm$0.012}  & \textbf{0.775$\pm$0.013}  \\
			BR & 0.345$\pm$0.018  & 0.783$\pm$0.040  &0.988$\pm$0.008  &0.750$\pm$0.036	 &0.388$\pm$0.016 			&0.753$\pm$0.016  &0.771$\pm$0.021  &0.754$\pm$0.013  \\
			ECC  & 0.442$\pm$0.014  &\textbf{0.789$\pm$0.036}  &0.991$\pm$0.008  &0.884$\pm$0.023   &0.557$\pm$0.015 &0.738$\pm$0.020  &0.811$\pm$0.012 &0.756$\pm$0.014  \\
			RAKEL   &0.329$\pm$0.016  &0.763$\pm$0.039  &0.993$\pm$0.006  &0.800$\pm$0.032   &0.456$\pm$0.019  	&0.735$\pm$0.017  &0.804$\pm$0.022  &0.720$\pm$0.014  \\
			LLSF  & 0.507$\pm$0.021  &0.716$\pm$0.035  &\textbf{0.997$\pm$0.005}  &0.912$\pm$0.015   & 0.682$\pm$0.028 & 0.790$\pm$0.014 &0.843$\pm$0.008 & 0.601$\pm$0.015  \\
			JFSC & 0.492$\pm$0.020  &0.691$\pm$0.040  &\textbf{0.997$\pm$0.004}  &0.908$\pm$0.016   &0.655$\pm$0.025  &0.779$\pm$0.011  &0.835$\pm$0.010  &0.746$\pm$0.012  \\
			\hline
			\multicolumn{1}{c}{\multirow{2}{*}{\begin{tabular}[c]{@{}c@{}}Comparing\\ algorithms\end{tabular}}} & \multicolumn{8}{c}{Macro-averaging F1$\uparrow$}\\ \cline{2-9}
			\multicolumn{1}{c}{}    & cal500 & emotions  & genbase  & medical   & enron  & image  & scene   & yeast \\
			\hline
			CAMEL &0.180$\pm$0.032  & \textbf{0.625$\pm$0.052}  	  &\textbf{0.971$\pm$0.030} & \textbf{0.779$\pm$0.043} & 0.325$\pm$0.044 & \textbf{0.660$\pm$0.030 }&\textbf{0.787$\pm$0.023} & \textbf{0.411$\pm$0.018}  \\
			BR &0.167$\pm$0.019  &0.620$\pm$0.044  	  &0.951$\pm$0.029 	   &0.640$\pm$0.060  		&0.236$\pm$0.016  		 &0.553$\pm$0.027  		   &0.623$\pm$0.026  		&0.391$\pm$0.021  \\
			ECC  &0.236$\pm$0.027  &0.622$\pm$0.043  	  &0.928$\pm$0.037  	   &0.755$\pm$0.054  		&0.303$\pm$0.030  		 &0.540$\pm$0.030  		   &0.662$\pm$0.026  		&0.395$\pm$0.015  \\
			RAKEL   &0.187$\pm$0.020  &0.614$\pm$0.044  	  &0.958$\pm$0.030  	   &0.689$\pm$0.051  		&0.256$\pm$0.017  		 &0.540$\pm$0.028  		   &0.644$\pm$0.024 		&0.381$\pm$0.020  \\
			LLSF  &0.180$\pm$0.031   &   0.615$\pm$0.056	  &\textbf{0.971$\pm$0.031} 	   	   &0.769$\pm$0.057 		&0.292$\pm$0.043  		 &0.554$\pm$0.031  		   &0.615$\pm$0.007 		&0.235$\pm$0.016  \\
			JFSC &\textbf{0.239$\pm$0.031} &0.345$\pm$0.023&\textbf{0.971$\pm$0.031}  	   &0.772$\pm$0.043  		&\textbf{0.339$\pm$0.048}  		 &0.559$\pm$0.035  		   &0.705$\pm$0.019  		&0.300$\pm$0.007  \\
			\hline
			\multicolumn{1}{c}{\multirow{2}{*}{\begin{tabular}[c]{@{}c@{}}Comparing\\ algorithms\end{tabular}}} & \multicolumn{8}{c}{Micro-averaging F1$\uparrow$}\\ \cline{2-9}
			\multicolumn{1}{c}{}    & cal500 & emotions  & genbase  & medical   & enron  & image  & scene   & yeast \\
			\hline
			CAMEL &0.337$\pm$0.017 &\textbf{0.649$\pm$0.041}  &0.988$\pm$0.012  &\textbf{0.835$\pm$0.019}  &\textbf{0.580$\pm$0.023}  &\textbf{0.659$\pm$0.031}  &\textbf{0.780$\pm$0.026} &\textbf{0.655$\pm$0.010} \\
			BR &0.339$\pm$0.016  		  &0.639$\pm$0.050  &0.978$\pm$0.014  &0.611$\pm$0.032  &0.359$\pm$0.015  &0.558$\pm$0.028  &0.619$\pm$0.023  &0.633$\pm$0.013  \\
			ECC &0.364$\pm$0.015  		  &0.642$\pm$0.046  &0.907$\pm$0.035  &0.796$\pm$0.023  &0.452$\pm$0.015  &0.541$\pm$0.030  &0.653$\pm$0.023  &0.643$\pm$0.017  \\
			RAKEL &0.351$\pm$0.018   &0.629$\pm$0.049  &0.983$\pm$0.011  &0.678$\pm$0.042  &0.392$\pm$0.014  &0.541$\pm$0.031 &0.629$\pm$0.026  &0.632$\pm$0.016  \\
			LLSF & 0.325$\pm$0.015  &0.637$\pm$0.049 & 0.992$\pm$0.003  &  0.823$\pm$0.027 & 0.534$\pm$0.025 &0.557$\pm$0.032  & 0.618$\pm$0.008 & 0.280$\pm$0.018  \\
			JFSC &\textbf{0.473$\pm$0.013}  	  &0.406$\pm$0.022  &\textbf{0.995$\pm$0.006}  &0.818$\pm$0.018  &0.555$\pm$0.026  &0.565$\pm$0.033  &0.695$\pm$0.022  &0.609$\pm$0.012  \\
			\hline
			\hline
		\end{tabular}
}
	\end{table*}
	\begin{table*}[!t]
		\centering
		\caption{Predictive performance of each algorithm (mean$\pm$std. deviation) on the large-scale datasets.}
		\label{largeResult}
        \resizebox{\textwidth}{85mm}{
		\begin{tabular}{ccccccccc}
			\hline
			\hline
			\multicolumn{1}{c}{\multirow{2}{*}{\begin{tabular}[c]{@{}c@{}}Comparing\\ algorithms\end{tabular}}} & \multicolumn{8}{c}{One-error$\downarrow$}\\ \cline{2-9}
			\multicolumn{1}{c}{} & science & arts & rcv1-s1 & rcv1-s2 & rcv1-s3 & rcv1-s4 & rcv1-s5 & business \\
			\hline
			CAMEL &\textbf{0.457$\pm$0.021}&0.462$\pm$0.024&\textbf{0.404$\pm$0.019}&\textbf{0.403$\pm$0.018}&\textbf{0.413$\pm$0.019}&\textbf{0.331$\pm$0.016}&0.404$\pm$0.010 & \textbf{0.101$\pm$0.009}\\
			BR &0.760$\pm$0.015  &0.642$\pm$0.022  &0.742$\pm$0.019  &0.723$\pm$0.024  &0.718$\pm$0.021  &0.662$\pm$0.021  &0.715$\pm$0.015 & 0.417$\pm$0.016\\
			ECC &0.574$\pm$0.022  &0.526$\pm$0.023  &0.471$\pm$0.020  &0.441$\pm$0.021  &0.448$\pm$0.021  &0.378$\pm$0.019  &0.425$\pm$0.016 & 0.153$\pm$0.008    \\
			RAKEL &0.623$\pm$0.014  &0.543$\pm$0.024  &0.613$\pm$0.019  &0.592$\pm$0.022  &0.578$\pm$0.020  &0.552$\pm$0.020  &0.575$\pm$0.014 & 0.201$\pm$0.009  \\
			LLSF & 0.486$\pm$0.013 & 0.454$\pm$0.027 &0.409$\pm$0.015  &  0.406$\pm$0.016 & 0.415$\pm$0.021 &0.333$\pm$0.016  &\textbf{0.399$\pm$0.018} & 0.122$\pm$0.016   \\
			JFSC &0.489$\pm$0.027  &\textbf{0.447$\pm$0.027}  &0.418$\pm$0.016  &0.407$\pm$0.014  &0.418$\pm$0.025  &0.337$\pm$0.015  &0.407$\pm$0.023    & 0.122$\pm$0.019\\
			\hline
			\multicolumn{1}{c}{\multirow{2}{*}{\begin{tabular}[c]{@{}c@{}}Comparing\\ algorithms\end{tabular}}} & \multicolumn{8}{c}{Hamming loss$\downarrow$}\\ \cline{2-9}
			\multicolumn{1}{c}{} & science & arts & rcv1-s1 & rcv1-s2 & rcv1-s3 & rcv1-s4 & rcv1-s5 & business \\
			\hline
			CAMEL &\textbf{0.030$\pm$0.001}  &0.055$\pm$0.002&\textbf{0.026$\pm$0.008}    &\textbf{0.023$\pm$0.001}  &\textbf{0.023$\pm$0.001}  & \textbf{0.018$\pm$0.001}  		  &\textbf{0.022$\pm$0.001}   & \textbf{0.024$\pm$0.001} \\
			BR &0.072$\pm$0.002 	 &0.079$\pm$0.003 	  &0.056$\pm$0.001 & 0.053$\pm$0.001  &0.053$\pm$0.001 & 0.041$\pm$0.001  		  &0.051$\pm$0.002  & 0.049$\pm$0.001 \\
		    ECC &0.036$\pm$0.002&0.063$\pm$0.002 & 0.028$\pm$0.001 & 0.024$\pm$0.001&0.024$\pm$0.001   &0.019$\pm$0.001 & 0.024$\pm$0.001  & 0.030$\pm$0.001  \\
			RAKEL &0.042$\pm$0.002   &0.075$\pm$0.002  	 &0.046$\pm$0.001  	   &0.039$\pm$0.001  		&0.035$\pm$0.001  	  	 &0.035$\pm$0.001  		  &0.036$\pm$0.003 & 0.035$\pm$0.002  \\
			LLSF &0.036$\pm$0.002  	 &\textbf{0.054$\pm$0.002} 	  &0.027$\pm$0.001  	  &0.025$\pm$0.001  	&0.025$\pm$0.001  	 &0.019$\pm$0.001  		  &0.023$\pm$0.001   & 0.048$\pm$0.007  \\
			JFSC &0.035$\pm$0.002  	 &0.057$\pm$0.002  		  &0.029$\pm$0.001  	   &0.025$\pm$0.001  		&0.025$\pm$0.001  		 &0.019$\pm$0.001  		  &0.025$\pm$0.001  & 0.027$\pm$0.002 \\
			\hline
			\multicolumn{1}{c}{\multirow{2}{*}{\begin{tabular}[c]{@{}c@{}}Comparing\\ algorithms\end{tabular}}} & \multicolumn{8}{c}{Coverage$\downarrow$}\\ \cline{2-9}
			\multicolumn{1}{c}{} & science & arts &  rcv1-s1 & rcv1-s2 & rcv1-s3 & rcv1-s4 & rcv1-s5 & business \\
			\hline
			CAMEL &\textbf{0.189$\pm$0.010}    &0.205$\pm$0.009   &0.151$\pm$0.008 	&\textbf{0.142$\pm$0.012}   &\textbf{0.131$\pm$0.006}    &0.143$\pm$0.003  	  &\textbf{0.132$\pm$0.005}  & \textbf{0.082$\pm$0.006} \\
			BR &0.303$\pm$0.011   &0.204$\pm$0.009   &0.393$\pm$0.011 & 0.341$\pm$0.013 & 0.351$\pm$0.018 & 0.294$\pm$0.015&0.336$\pm$0.013 & 0.141$\pm$0.002 \\
			ECC  &0.196$\pm$0.009 &0.229$\pm$0.009   &0.166$\pm$0.011  		&0.154$\pm$0.007  		 &0.154$\pm$0.008  		  &0.108$\pm$0.003  	   &0.145$\pm$0.001 & 0.104$\pm$0.001\\
			RAKEL   &0.209$\pm$0.012  	 &0.214$\pm$0.008  	  &0.273$\pm$0.011  	&0.329$\pm$0.012  	&0.293$\pm$0.017  	  &0.273$\pm$0.012     &0.246$\pm$0.012   & 0.107$\pm$0.003\\
			LLSF  &0.197$\pm$0.014   &\textbf{0.195$\pm$0.011} &0.141$\pm$0.009  &0.146$\pm$0.008 	 &0.133$\pm$0.008  	  &0.109$\pm$0.006 	    &0.133$\pm$0.006  & 0.086$\pm$0.013 \\
			JFSC &0.196$\pm$0.011  	  &0.233$\pm$0.018    &\textbf{0.140$\pm$0.006} 	&0.143$\pm$0.009  	 &0.136$\pm$0.010  	  &\textbf{0.106$\pm$0.005}   &0.139$\pm$0.006  & 0.086$\pm$0.011  \\
			\hline
			\multicolumn{1}{c}{\multirow{2}{*}{\begin{tabular}[c]{@{}c@{}}Comparing\\ algorithms\end{tabular}}} & \multicolumn{8}{c}{Ranking loss$\downarrow$}\\ \cline{2-9}
			\multicolumn{1}{c}{} & science & arts &  rcv1-s1 & rcv1-s2 & rcv1-s3 & rcv1-s4 & rcv1-s5 & business \\
			\hline
			CAMEL &\textbf{0.139$\pm$0.007}  &\textbf{0.135$\pm$0.008} &\textbf{0.058$\pm$0.003}    &0.077$\pm$0.005    &\textbf{0.047$\pm$0.003}  &0.057$\pm$0.002 	 &0.073$\pm$0.002 & \textbf{0.040$\pm$0.004}\\
			BR &0.245$\pm$0.009 & 0.145$\pm$0.006 & 0.197$\pm$0.006 &0.190$\pm$0.008 &0.198$\pm$0.010&0.173$\pm$0.009 &  0.181$\pm$0.006 & 0.088$\pm$0.006 \\
			ECC  &0.151$\pm$0.006  		&0.164$\pm$0.007  	 &0.074$\pm$0.005  		  &0.069$\pm$0.003  &0.070$\pm$0.002  	&0.047$\pm$0.004  	 &0.063$\pm$0.003  & 0.055$\pm$0.002  \\
			RAKEL   &0.195$\pm$0.007  		&0.156$\pm$0.008  	 &0.183$\pm$0.006 	  &0.153$\pm$0.008     &0.178$\pm$0.010  	&0.112$\pm$0.009  	 &0.123$\pm$0.006   & 0.067$\pm$0.005 \\
			LLSF  &0.149$\pm$0.009  		&0.141$\pm$0.009  	 &0.060$\pm$0.003 		  &\textbf{0.060$\pm$0.004}  	   &0.048$\pm$0.003  		&\textbf{0.034$\pm$0.003} 		 &\textbf{0.045$\pm$0.003}  &0.045$\pm$0.009  \\
			JFSC &0.147$\pm$0.008  		&0.159$\pm$0.009  			 &0.061$\pm$0.003  		  &0.062$\pm$0.006 	   &0.061$\pm$0.004  		&0.047$\pm$0.003  		 &0.060$\pm$0.003   & 0.045$\pm$0.008 \\
			\hline
			\multicolumn{1}{c}{\multirow{2}{*}{\begin{tabular}[c]{@{}c@{}}Comparing\\ algorithms\end{tabular}}} & \multicolumn{8}{c}{Average precision$\uparrow$}\\ \cline{2-9}
			\multicolumn{1}{c}{} & science & arts & rcv1-s1 & rcv1-s2 & rcv1-s3 & rcv1-s4 & rcv1-s5 & business \\
			\hline
			CAMEL &\textbf{0.624$\pm$0.016}&0.607$\pm$0.018  &0.615$\pm$0.009  &\textbf{0.644$\pm$0.012 } &\textbf{0.635$\pm$0.010} &\textbf{0.717$\pm$0.008}  &\textbf{0.626$\pm$0.009 }  & \textbf{0.891$\pm$0.009} \\
			BR &0.383$\pm$0.011  	  &0.514$\pm$0.013  & 0.353$\pm$0.011 & 0.382$\pm$0.015  &0.382$\pm$0.015  & 0.443$\pm$0.013 & 0.390$\pm$0.009 & 0.709$\pm$0.008   \\
			ECC  &0.516$\pm$0.020  		  &0.553$\pm$0.018  &0.545$\pm$0.016  &0.587$\pm$0.015  &0.585$\pm$0.016  &0.677$\pm$0.017  &0.600$\pm$0.009  & 0.844$\pm$0.005  \\
			RAKEL   &0.487$\pm$0.012  	 &0.526$\pm$0.015 &0.424$\pm$0.012  &0.489$\pm$0.016  &0.459$\pm$0.014  &0.479$\pm$0.012  &0.432$\pm$0.009 & 0.858$\pm$0.007  \\
			LLSF  & 0.594$\pm$0.021  &\textbf{0.631$\pm$0.016} & \textbf{0.627$\pm$0.009} & 0.637$\pm$0.008  & 0.632$\pm$0.013 & 0.714$\pm$0.010  & 0.625$\pm$0.013  & 0.867$\pm$0.013\\
			JFSC &0.595$\pm$0.020  		  &0.621$\pm$0.020  &0.606$\pm$0.008  &0.630$\pm$0.009  &0.624$\pm$0.014  &0.700$\pm$0.012  &0.624$\pm$0.013 & 0.874$\pm$0.018  \\
			\hline
			\multicolumn{1}{c}{\multirow{2}{*}{\begin{tabular}[c]{@{}c@{}}Comparing\\ algorithms\end{tabular}}} & \multicolumn{8}{c}{Macro-averaging F1$\uparrow$}\\ \cline{2-9}
			\multicolumn{1}{c}{} & science & arts & rcv1-s1 & rcv1-s2 & rcv1-s3 & rcv1-s4 & rcv1-s5 & business \\
			\hline
			CAMEL &\textbf{0.310$\pm$0.038} &\textbf{0.312$\pm$0.029} &0.250$\pm$0.023 &\textbf{0.258$\pm$0.022} & 0.247$\pm$0.025 &\textbf{0.340$\pm$0.031}  &0.253$\pm$0.016 & \textbf{0.326$\pm$0.046} \\
			BR &0.215$\pm$0.048  &0.257$\pm$0.020  &0.232$\pm$0.018  		&0.210$\pm$0.017  		 &0.221$\pm$0.019  		  &0.313$\pm$0.016  &0.236$\pm$0.019  & 0.249$\pm$0.017  \\
			ECC &0.285$\pm$0.024  &0.282$\pm$0.021  		&0.271$\pm$0.023  		&0.257$\pm$0.022  		 &0.266$\pm$0.012  		  &0.334$\pm$0.018  &\textbf{0.285$\pm$0.014} & \textbf{0.326$\pm$0.032}\\
			RAKEL &0.267$\pm$0.028  &0.275$\pm$0.019  	    &0.266$\pm$0.019  		&0.237$\pm$0.023  		 &0.243$\pm$0.017  		  &0.322$\pm$0.017  &0.255$\pm$0.018  & 0.307$\pm$0.024 \\
			LLSF &0.312$\pm$0.038  &0.219$\pm$0.032  &0.261$\pm$0.022	&0.257$\pm$0.025  		 &\textbf{0.270$\pm$0.027} 	  &0.334$\pm$0.031 &  0.217$\pm$0.018 & 0.325$\pm$0.028   \\
			JFSC &0.308$\pm$0.039  &0.305$\pm$0.032 &\textbf{0.308$\pm$0.026}  	&0.249$\pm$0.019  	 &0.258$\pm$0.024   &0.337$\pm$0.032 &0.254$\pm$0.019 & 0.318$\pm$0.036\\
			\hline
			\multicolumn{1}{c}{\multirow{2}{*}{\begin{tabular}[c]{@{}c@{}}Comparing\\ algorithms\end{tabular}}} & \multicolumn{8}{c}{Micro-averaging F1$\uparrow$}\\ \cline{2-9}
			\multicolumn{1}{c}{} & science & arts &  rcv1-s1 & rcv1-s2 & rcv1-s3 & rcv1-s4 & rcv1-s5 & business \\
			\hline
			CAMEL &0.428$\pm$0.018  &0.415$\pm$0.015 &0.401$\pm$0.015  &\textbf{0.437$\pm$0.017}  &\textbf{0.431$\pm$0.025}  &\textbf{0.491$\pm$0.017} &\textbf{0.441$\pm$0.015} & \textbf{0.746$\pm$0.011}\\
			BR &0.277$\pm$0.013  	&0.349$\pm$0.018 			&0.301$\pm$0.009  &0.310$\pm$0.009  &0.307$\pm$0.013  &0.356$\pm$0.009  &0.321$\pm$0.009 & 0.595$\pm$0.003   \\
			ECC  &0.343$\pm$0.028  		&0.377$\pm$0.018  			&0.385$\pm$0.016  &0.410$\pm$0.022  &0.414$\pm$0.013  &0.482$\pm$0.024  &0.440$\pm$0.011  & 0.690$\pm$0.007 \\
			RAKEL   &0.337$\pm$0.014  		&0.368$\pm$0.017  		&0.341$\pm$0.010  &0.337$\pm$0.008  &0.335$\pm$0.014  &0.369$\pm$0.008  &0.350$\pm$0.008  & 0.701$\pm$0.014 \\
			LLSF  &0.446$\pm$0.025  	&0.368$\pm$0.018  	&\textbf{0.463$\pm$0.016}  & 0.432$\pm$0.018 &0.428$\pm$0.023  & 0.478$\pm$0.017  &0.438$\pm$0.019 & 0.693$\pm$0.035\\
			JFSC &\textbf{0.449$\pm$0.026}  	&\textbf{0.442$\pm$0.017} 	&0.456$\pm$0.008  &0.422$\pm$0.011  &0.424$\pm$0.012  &0.482$\pm$0.013  &0.438$\pm$0.011 & 0.712$\pm$0.021 \\
			\hline
			\hline
		\end{tabular}
    }
	\end{table*}
\begin{figure*}[!t]
\subfigure[$\alpha$]{
    \begin{minipage}[b]{0.24\textwidth}
      \centering
      \includegraphics[width=1.7in]{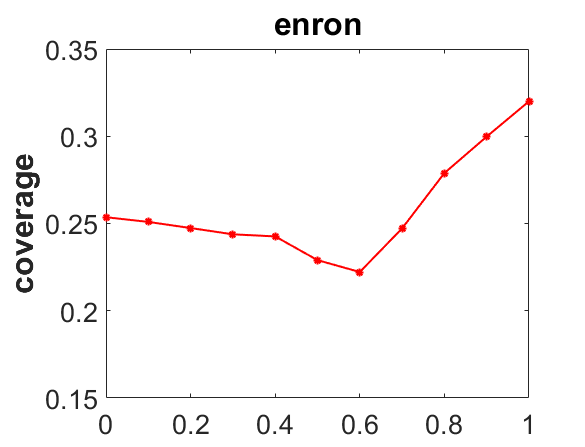}
    \end{minipage}}%
  \subfigure[$\lambda_1$]{
    \begin{minipage}[b]{0.24\textwidth}
      \centering
      \includegraphics[width=1.7in]{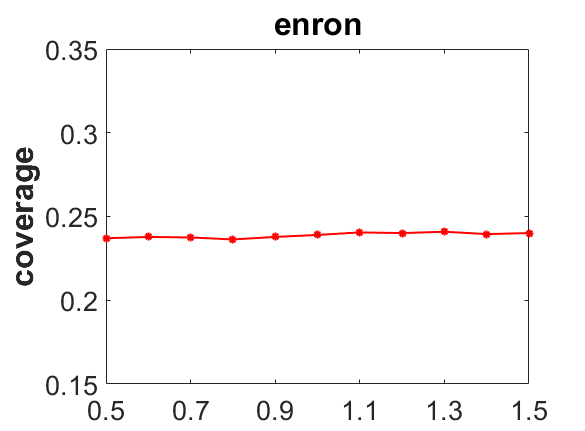}
    \end{minipage}}
      \subfigure[$\lambda_2$]{
    \begin{minipage}[b]{0.24\textwidth}
      \centering
      \includegraphics[width=1.7in]{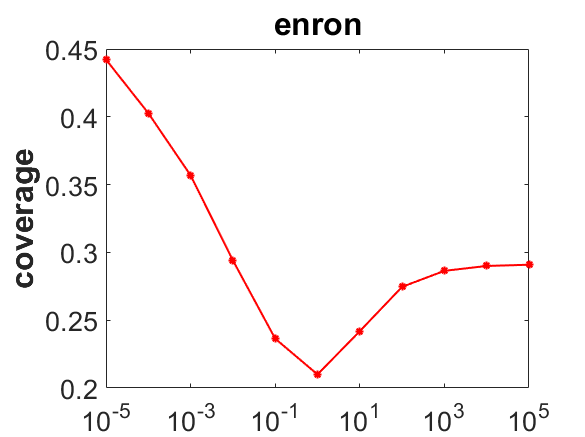}
    \end{minipage}}
      \subfigure[Convergence Curve]{
    \begin{minipage}[b]{0.24\textwidth}
      \centering
      \includegraphics[width=1.7in]{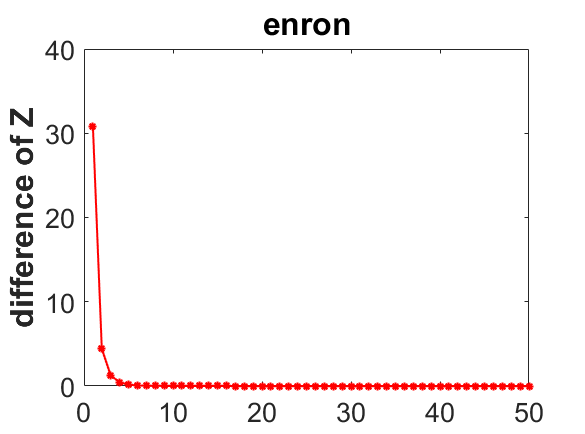}
    \end{minipage}}
  \caption*{Figure 1: Parameter sensitivity and convergence analysis of CAMEL on the enron dataset.}
  \label{fig2} 
\end{figure*}
Table 2 and 3 report the detailed experimental results on the regular-scale and large-scale datasets respectively, where the best performance among all the algorithms is shown in boldface. From the two result tables, we can see that CAMEL outperforms other comparing algorithms in most cases. Specifically, on the regular-size datasets (Table 2), across all the evaluation metrics, CAMEL ranks first in 80.4\% (45/56) cases, and on the large-scale datasets (Table 3), across all the evaluation metrics, CAMEL ranks first in 69.6\% (39/56) cases. Compared with the three well-established algorithms BR, ECC, and RAKEL, CAMEL introduces a new type of label correlations, i.e., collaborative relationships among labels, and achieves superior performance in 93.8\% (315/336) cases. Compared with the two state-of-the-art algorithms LLSF and JFSC, instead of employing simple similarity measures to regularize the hypothesis space, CAMEL introduces a novel method to learn label correlations for explicitly correlating the final predictions, and achieves superior performance in 80.4\% (180/224) cases. These comparative results clearly demonstrate the effectiveness of the collaboration based multi-label learning approach.
\subsubsection{Sensitivity Analysis}
In this section, we first investigate the sensitivity of CAMEL with respect to the two tradeoff parameters $\lambda_1$ and $\lambda_2$, and the parameter $\alpha$ that controls the degree of collaboration, then illustrate the convergence of CAMEL. Due to page limit, we only report the experimental results on the enron dataset using the \textsl{Coverage} ($\downarrow$) metric. Concretely, we study the performance of CAMEL when we vary one parameter while keeping other parameters fixed at their best setting. Figure 1(a), 1(b), and 1(c) show the sensitivity curve of CAMEL with respect to $\alpha$, $\lambda_1$, and $\lambda_2$ respectively. It can be seen that $\alpha$ and $\lambda_2$ have an important influence on the final performance, because $\alpha$ and $\lambda_2$ control the collaboration degree and the model complexity.
Figure 1(d) illustrates the convergence of CAMEL by using the difference of the optimization variable $\mathbf{Z}$ between two successive iterations, i.e., $\Delta\mathbf{Z}=\left\|\mathbf{Z}^{(t)} - \mathbf{Z}^{(t-1)}\right\|_F$. From Figure 1(d), we can observe that $\Delta\mathbf{Z}$ quickly decreases to 0 within a few number of iterations. Hence the convergence of CAMEL is demonstrated.

\section{Conclusion}
In this paper, we make a key assumption for multi-label learning that \textsl{for each individual label, the final prediction involves the collaboration between its own prediction and the predictions of other labels.} Guided by this assumption, we propose a novel method to learn the high-order label correlations via sparse reconstruction in the label space. Besides, by seamlessly integrating the learned label correlations into model training, we propose a novel multi-label learning approach that aims to explicitly account for
the correlated predictions of labels while training the desired model simultaneously. Extensive experimental results show that our approach outperforms the state-of-the-art counterparts.

Despite the demonstrated effectiveness of CAMEL, it only considers the global collaborative relationships between labels, by assuming that such collaborative relationships are shared by all the instances. However, as different instances have different characteristics, such collaborative relationships may be shared by only a subset of instances rather than all the instances. Therefore, our further work is to explore different collaborative relationships between labels for different subsets of instances.
\section{Acknowledgements}
This work was supported by MOE, NRF, and NTU.
\section*{Appendix A. The ADMM Procedure}
To solve problem~(\ref{eq3}) by ADMM, we first reformulate problem~(\ref{eq3}) into the following equivalent form:
\begin{gather}
\label{eqcons}
\min_{\mathbf{S}_j,\mathbf{z}}\frac{1}{2}\left\|\mathbf{Y}_{-j}\mathbf{S}_i-\mathbf{Y}_j\right\|_2^2+\lambda\left\|\mathbf{z}\right\|_1\\
\nonumber
\text{s.t.}\quad \mathbf{S}_j - \mathbf{z} = 0
\end{gather}
Following the ADMM procedure, the above constrained optimization problem~(\ref{eqcons}) can be solved as a series of unconstrained minimization problems using augmented Lagrangian function, which is presented as:
\begin{gather}
\label{eqLag}
\mathcal{L}(\mathbf{S}_j,\mathbf{z},\boldsymbol{\mu}) = \frac{1}{2}\left\|\mathbf{Y}_{-j}\mathbf{S}_j-\mathbf{Y}_j\right\|_2^2+\lambda\left\|\mathbf{z}\right\|_1+\\
\nonumber
\mathbf{v}^\top(\mathbf{S}_j - \mathbf{z})+\frac{\rho}{2}\left\|\mathbf{S}_j-\mathbf{z}\right\|_2^2
\end{gather}
Here, $\rho$ is the penalty parameter and $\mathbf{v}$ is the Lagrange multiplier. By introducing the scaled dual variable $\boldsymbol{\mu}=\frac{1}{\rho}\mathbf{v}$,
a sequential minimization of the scaled ADMM iterations can be conducted by updating the three variables $\mathbf{S}_j$, $\mathbf{z}$ and $\boldsymbol{\mu}$ sequentially:
\begin{align}
\nonumber
\mathbf{S}_j^{(k+1)} &= (\mathbf{Y}_{-j}^\top\mathbf{Y}_{-j} + \rho\mathbf{I})^{-1}(\mathbf{Y}_{-j}^\top\mathbf{Y}_j+\rho(\mathbf{z}^{(k)} - \boldsymbol{\mu}^{(k)}))\\
\nonumber
\mathbf{z}^{(k+1)} &= S_{\lambda/\rho}(\mathbf{S}_j^{(k+1)}+\boldsymbol{\mu}^{(k)})\\
\boldsymbol{\mu}^{(k+1)} &= \boldsymbol{\mu}^{(k)}+\mathbf{S}^{(k+1)} - \mathbf{z}^{(k+1)}
\end{align}
where $S$ is the proximity operator of the $\ell_1$ norm, which is defined as $S_{\omega}(a) = (a-\omega)_{+}-(-a-\omega)_{+}$.
\section*{Appendix B. Model Parameter Optimization}
The Lagrangian of problem~(\ref{updateW}) is expressed as:
\begin{gather}
\mathcal{L}({\mathbf{W},\mathbf{E},\mathbf{A},\mathbf{b}}) = \tr(\mathbf{E}^\top\mathbf{E})+\lambda_2\tr(\mathbf{W}^\top\mathbf{W})+\\
\nonumber
\tr{(\mathbf{A}^\top(\mathbf{Z}-\phi(\mathbf{X})\mathbf{W}-\mathbf{1}\mathbf{b}^\top-\mathbf{E}))}
\end{gather}
where $\tr$ is the trace operator, and $\mathbf{A} = [\mathbf{a}_1,\mathbf{a}_2,\cdots,\mathbf{a}_n]^\top\in\mathbb{R}^{n\times q}$ is the introduced matrix that stores the Lagrangian multipliers. Besides, we have used the property of trace operator that $\tr(\mathbf{W}^\top\mathbf{W}) = \left\|\mathbf{W}\right\|_F^2$. By Setting the gradient w.r.t. $\mathbf{E},\mathbf{A},\mathbf{W},\mathbf{b}$ to 0 respectively, the following equations will be induced:
\begin{align}
\nonumber
	\frac{\partial\mathcal{L}}{\partial\mathbf{E}} = 0&\Rightarrow \mathbf{A} = \mathbf{E}\\
\nonumber
	\frac{\partial\mathcal{L}}{\partial\mathbf{A}} = 0&\Rightarrow \mathbf{Z}-\phi(\mathbf{X})\mathbf{W}-\mathbf{1}\mathbf{b}^\top=\mathbf{E}\\
\nonumber
	\frac{\partial\mathcal{L}}{\partial\mathbf{W}} = 0&\Rightarrow \mathbf{W} = \frac{1}{\lambda_2}\phi(\mathbf{X})^\top\mathbf{A}\\
	\frac{\partial\mathcal{L}}{\partial\mathbf{b}} = 0&\Rightarrow \mathbf{A}^\top\mathbf{1} = \mathbf{0}
\end{align}
The above linear equations can be simplified by the following steps:
\begin{align}
\nonumber
\mathbf{Z}&=\phi(\mathbf{X})\mathbf{W}+\mathbf{1}\mathbf{b}^\top+\mathbf{E}\\
\nonumber
\mathbf{Z}&=\frac{1}{\lambda_2}\phi(\mathbf{X})\phi(\mathbf{X})^\top\mathbf{A} + \mathbf{1}\mathbf{b}^\top+\mathbf{A}\\
\mathbf{Z}&=\frac{1}{\lambda_2}\mathbf{K}\mathbf{A} + \mathbf{1}\mathbf{b}^\top+\mathbf{A}
\end{align}
Here, we define $\mathbf{H} = \frac{1}{\lambda_2}\mathbf{K}+\mathbf{I}$, then we can obtain:
\begin{align}
\nonumber
\mathbf{H}\mathbf{A}+\mathbf{1}\mathbf{b}^\top &= \mathbf{Z}\\
\nonumber
\mathbf{A}+\mathbf{H}^{-1}\mathbf{1}\mathbf{b}^\top &= \mathbf{H}^{-1}\mathbf{Z}\\
\nonumber
\mathbf{1}^\top\mathbf{H}^{-1}\mathbf{1}\mathbf{b}^\top &= \mathbf{1}^\top\mathbf{H}^{-1}\mathbf{Z}\\
\mathbf{b}^\top &= \frac{\mathbf{1}\mathbf{H}^{-1}\mathbf{Z}}{\mathbf{1}^\top\mathbf{H}^{-1}\mathbf{1}}
\end{align}
In this way, $\mathbf{A}$ can be calculated by $\mathbf{A} = \mathbf{H}^{-1}(\mathbf{Z}-\mathbf{1}\mathbf{b}^\top)$.
\bibliographystyle{aaai}
\bibliography{aaai19}

\end{document}